\documentclass[sigconf, screen]{acmart}

\AtBeginDocument{%
  }

\usepackage{multirow}
\usepackage{makecell}
\usepackage{xcolor}   
\usepackage{colortbl} 
\usepackage{arydshln} 
\usepackage{pifont}
\usepackage{bbm}
\usepackage{dsfont}

\definecolor{titlegray}{HTML}{F2F2F2} 
\definecolor{ourspink}{HTML}{FFE1E5}
\definecolor{venuegray}{HTML}{B0B0B0} 

\renewcommand\footnotetextcopyrightpermission[1]{}

\setcopyright{none}
\begin{document}

\title{ExACT: Exemplar-Driven Calibrated Refinement for Training-Free Visual Grounding in Remote Sensing Images}

\author{Zixiao Zhang}
\affiliation{%
  \institution{Xidian University}
  \country{}
}
\email{zhangzx1999@stu.xidian.edu.cn}

\author{Lingling Li}
\affiliation{%
  \institution{Xidian University}
  \country{}
}
\email{llli@xidian.edu.cn}

\author{Pei He}
\affiliation{%
  \institution{Xidian University}
  \country{}
}
\email{hepei@stu.xidian.edu.cn}

\author{Xu Liu}
\affiliation{%
  \institution{Xidian University}
  \country{}
}
\email{xuliu361@163.com}

\author{Licheng Jiao}
\affiliation{%
  \institution{Xidian University}
  \country{}
}
\email{lchjiao@mail.xidian.edu.cn}

\renewcommand{\shortauthors}{Zhang et al.}

\begin{abstract}
  Remote sensing visual grounding (RSVG) aims to locate specific objects in high-resolution RS imagery using free-form natural language descriptions. While recent advances in multimodal large language models (MLLMs) show great potential for such open-vocabulary RSVG, their training-free adaptation is hindered by the modality gap between abstract linguistic semantics and fine-grained visual cues. In cluttered RS scenes, this gap inevitably causes severe localization drift. To bridge this gap, we propose \textbf{Ex}empl\textbf{a}r-driven \textbf{C}alibrated Refinemen\textbf{t} (\textbf{ExACT}), a novel training-free framework driven by a one-shot visual prompting mechanism to explicitly provide discriminative structural guidance for precise pixel-level localization. Specifically, we propose a Vision Exemplar-based Calibrator (VEC) that extracts fine-grained visual correspondences from the given exemplar to rectify the rough cross-modal priors from frozen MLLMs, effectively suppressing background artifacts and accurately outlining target boundaries. Subsequently, a Structure-Aware Refiner (SAR) employs an iterative merge-and-select clustering strategy to consolidate the calibrated priors into high-quality positive and negative geometric prompts. These prompts then guide the Segment Anything Model (SAM) to achieve precise pixel-level predictions. Extensive experiments confirm the superiority of ExACT over existing training-free and weakly-supervised methods. All the source codes will be made publicly available.
\end{abstract}




\maketitle

\section{Introduction}

\begin{figure}
  \centering
  \includegraphics[width=\linewidth]{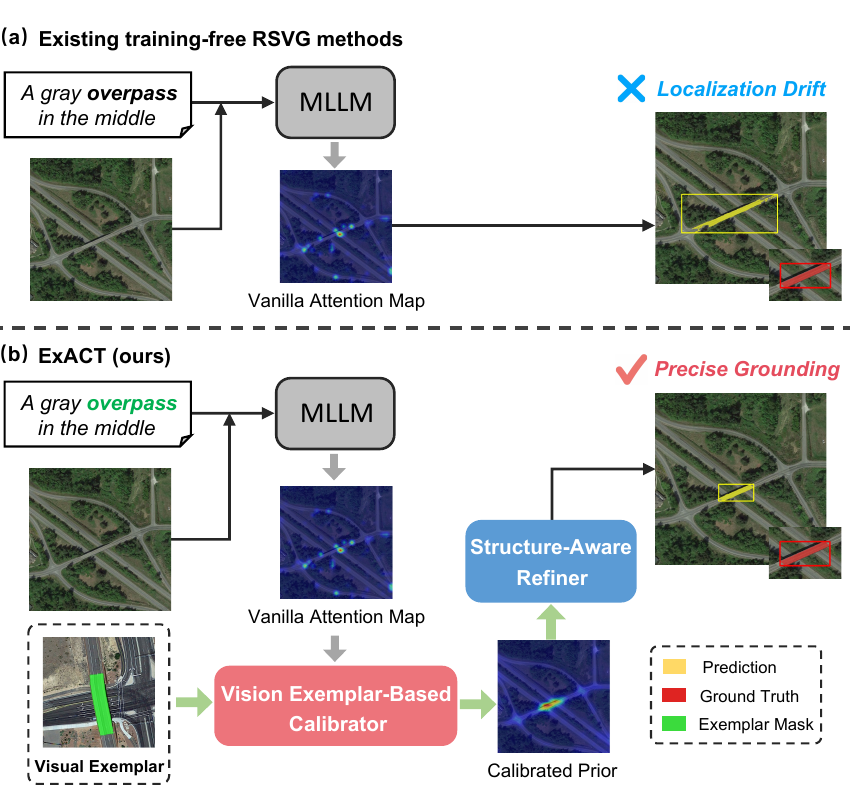}
  \caption{The motivation of our ExACT. (a) Relying solely on linguistic queries, current training-free methods lack explicit visual constraints, leading to severe localization drift. (b) ExACT introduces a visual exemplar for discriminative structural guidance. The VEC rectifies noisy attention maps into calibrated priors, which the SAR then consolidates to achieve precise pixel-level grounding.}
  \Description{}
  \label{motivation}
\end{figure}

Remote sensing visual grounding (RSVG) aims to locate queried objects in high-resolution RS imagery based on natural language descriptions \cite{vg-survey}. This advanced task allows users to interactively explore geographic structures and spatial relationships in RS scenes, playing a vital role in applications such as ecological protection\cite{agriculture}, disaster surveillance\cite{disaster}, and urban planning\cite{urban}. Early studies\cite{rsvg1,rsvg,LGFormer,RMSIN,chai2026like} employed fully supervised methods to learn visual-linguistic alignment and produce box- or pixel-level predictions. However, these methods heavily depend on labor-intensive annotated datasets. Furthermore, their closed-set nature limits their ability to generalize to out-of-distribution objects. To move beyond fixed categories and enable open-vocabulary perception, recent research has employed multimodal large language models (MLLMs)\cite{qwen2} to visually ground language queries \cite{heads,f-lmm,lisa}.

 Leveraging extensive world knowledge and strong instruction-following abilities of MLLMs, recent RSVG models\cite{lisa,groundvlp,geochat,geoground,lhrs,rsvg-zero} have become a leading paradigm, generally divided into two streams: supervised training-based and training-free inference. The first stream focuses on RS-specific grounding MLLMs, which typically treat visual grounding as a next-token-prediction task, either representing bounding box coordinates as discrete text tokens or predicting special tokens to encode segmentation masks \cite{lisa,geochat,geoground,lhrs,pang2025vhm}. However, to perform well, they rely heavily on extensive supervised fine-tuning or multitask instruction tuning to adapt generalist models to the RS domain. Conversely, to avoid training overhead, the second stream explores training-free inference methods pioneered by RSVG-ZeroOV \cite{rsvg-zero}. This approach uses cross-attention maps from frozen MLLMs and incorporates self-attention maps from off-the-shelf diffusion models (DMs) \cite{stable-diffusion} to enhance the structural details required for pixel-level localization without modifying any parameters.

Despite the elegance of existing training-free methods, they still face a critical bottleneck: relying on the textual query as the only signal. Linguistic cues naturally encode sparse information, often failing to convey the fine-grained visual details necessary to delineate precise object boundaries or distinguish subtle appearance differences. For example, ``a gray overpass in the middle” cannot help the model understand the physical boundary of ``overpass” or clarify the implicit differences between ``overpass” and ``roads”.  In cluttered RS scenes, this  modality gap inevitably triggers \textbf{spatio-semantic misalignment}, ultimately leading to severe localization drift. As shown in Figure. \ref{motivation}(a), this misalignment manifests in the language-guided localization priors (i.e., vanilla attention maps from MLLMs) through two notable issues: \textbf{\textit{coordinate-driven bias}} and \textbf{\textit{high-activation artifacts}}. Driven by coordinate-based regression objectives, the textual attention tends to overfit to bounding-box edges rather than to the object's actual physical extent. Furthermore, due to unconstrained global-context reasoning, high activations are frequently misallocated to irrelevant backgrounds. While recent training-free methods employ structural priors from off-the-shelf DMs to recover object shapes, they fail to resolve this fundamental flaw. As DMs are inherently generative and prioritize low-level spatial continuity, their unconstrained priors lack discriminative semantic awareness. Without explicit visual constraints, the model struggles to disentangle the target from visually similar surroundings. For instance, querying a ``gray overpass'' often causes activations to incorrectly extend into connected roads, leading to severe \textbf{\textit{boundary leakage}}.  Ultimately, relying solely on language prompts leads to semantic ambiguity and geometric uncertainty, especially in complex geospatial scenes.

Motivated by the insight that ``a picture is worth a thousand words'' \cite {gu2024context}, we investigate how visual exemplars can provide the discriminative structural priors that purely text-guided models lack. Specifically, leveraging the powerful feature representations of vision foundation models (VFMs) such as DINOv3\cite{dino}, we can establish dense visual correspondences by matching all patch-level features of the target image against the prompted foreground features of a given exemplar image. 
Compared with the coarse text-driven attention, these fine-grained visual similarities capture rich semantic details crucial for delineating object boundaries. Despite this advantage, pure visual matching lacks language-guided referring capability. In scenes with multiple instances of the same category (e.g., two cars), it activates all candidates equally, making it difficult to identify the exact target described in the query.
Therefore, it is evident that language-guided localization cues from MLLMs and cross-image visual correspondences from VFMs are naturally complementary. The former provides the essential semantic instruction for isolating the specific target, while the latter supplies the fine-grained visual matching required for boundary certainty.

Accordingly, we propose \textbf{ExACT} (\textbf{Ex}empl\textbf{a}r-driven \textbf{C}alibrated Refinemen\textbf{t}), a novel training-free framework that explicitly leverages a one-shot visual exemplar to supplement coarse MLLM priors with fine-grained structural guidance. By employing this exemplar as a cross-modal visual prompt, our approach effectively bridges the visual-semantic gap inherent in purely text-prompted paradigms. As illustrated in Figure \ref{motivation}(b), ExACT achieves this through two core modules: a \textbf{Vision Exemplar-based Calibrator (VEC)} and a \textbf{Structure-Aware Refiner (SAR)}. First, VEC extracts dense feature-level correlations from the visual prompt using VFMs to rectify the noisy textual localization priors from frozen MLLMs, effectively suppressing high-activation background artifacts and boundary leakage. Subsequently, the SAR addresses geometric uncertainty and reduces noisy proposals. Operating as a coarse-to-fine \textit{merge-and-select} pipeline, it further filters spurious activations to extract high-quality geometric prompts for the foundation segmentation model SAM, thereby ensuring precise pixel-level localization.
The contributions of our work are summarized as follows:

\begin{itemize}
\item \textbf{\textit{A Novel Exemplar-Driven Framework:}} We propose ExACT, a training-free RSVG framework driven by a one-shot visual prompting mechanism. By coupling MLLM semantic reasoning with the fine-grained visual correspondences of VFM, it effectively bridges the visual-semantic gap inherent to purely text-prompted paradigms.

\item \textbf{\textit{Explicit Calibration and Refinement:}} To address spatio-semantic misalignment, we introduce a Vision Exemplar-Based Calibrator (VEC) that leverages dense feature-level correlations to explicitly rectify noisy MLLM priors, and a Structure-Aware Refiner (SAR) that distills these calibrated priors into robust pixel-level masks.
\item \textbf{\textit{SOTA Training-Free Performance:}} Extensive experiments on two RSVG benchmarks demonstrate that ExACT significantly outperforms existing MLLM grounding models without any parameter updates, confirming the superiority of our exemplar-driven paradigm.

\end{itemize}

\section{Related Works}
\noindent\textbf{Remote Sensing Visual Grounding.}
RSVG encompasses two typical tasks: Remote sensing Referring Expression Comprehension (RSREC) and Segmentation (RSRES) \cite{RRSECS}. Although fully supervised methods \cite{rsvg1,rsvg,LPVA,LGFormer,RMSIN} deliver high accuracy, they rely heavily on labor-intensive annotations and struggle to generalize to out-of-distribution attributes.
To achieve open-vocabulary perception, recent MLLM-based approaches \cite{lisa,geochat,lhrs,pang2025vhm,rsvg-zero} generally fall into two streams: supervised training-based and training-free inference. The first stream develops RS-specific grounding MLLMs \cite{lisa,geochat,lhrs,pang2025vhm} via supervised fine-tuning \cite{vlm} or multitask instruction tuning \cite{instruction} to bridge the domain gap. However, these methods require costly retraining, and their next-token-prediction paradigms inherently limit the continuous spatial precision. 

Training-free inference has emerged as the second stream that seeks to alleviate these high training costs. For instance, RSVG-ZeroOV and its extension \cite{rsvg-zero,rsvg-zero-video} to RS images and videos extract localization cues from frozen MLLMs and combine structural priors from DMs to recover spatial details. Nevertheless, lacking explicit visual guidance, their purely text-driven inference inevitably suffers from severe localization shifts in cluttered RS scenes. By introducing a one-shot visual exemplar, ExACT supplies the discriminative structural guidance missing in purely text-prompted paradigms, effectively resolving spatio-semantic misalignment without any parameter updates.

\noindent\textbf{Visual Reference Prompting.}
Visual reference prompting is widely applied in image-prompted segmentation \cite{perSAM,IPSeg,Matcher,Bridge,synpo}, where visual exemplars explicitly specify the content that users intend to segment. These methods usually generate geometric prompts for foundation segmentation models by pixel-level feature matching. For instance, IPSeg \cite{IPSeg} selects representative points from dense correlation maps for training-free open-world segmentation.
To address ambiguity in a single modality, hybrid frameworks such as VLP-SAM \cite{VLP-SAM} combine visual references and textual labels.

However, adapting these paradigms to the RS domain remains challenging. Pure visual matching relies heavily on low-level correlations and struggles with high-level semantic reasoning. Although hybrid prompting models incorporate text labels to mitigate this, they are tailored for natural images and still demand parameter optimization to align unified prompt embeddings. To overcome this, ExACT offers a completely training-free framework for complex RS scenes. Instead of tuning embeddings, ExACT uniquely introduces a one-shot visual exemplar to supply discriminative structural guidance. This plug-and-play approach effectively bridges the visual-semantic gap, ensuring precise pixel-level localization with zero parameter overhead.

\begin{figure}
  \centering
  \includegraphics[width=\linewidth]{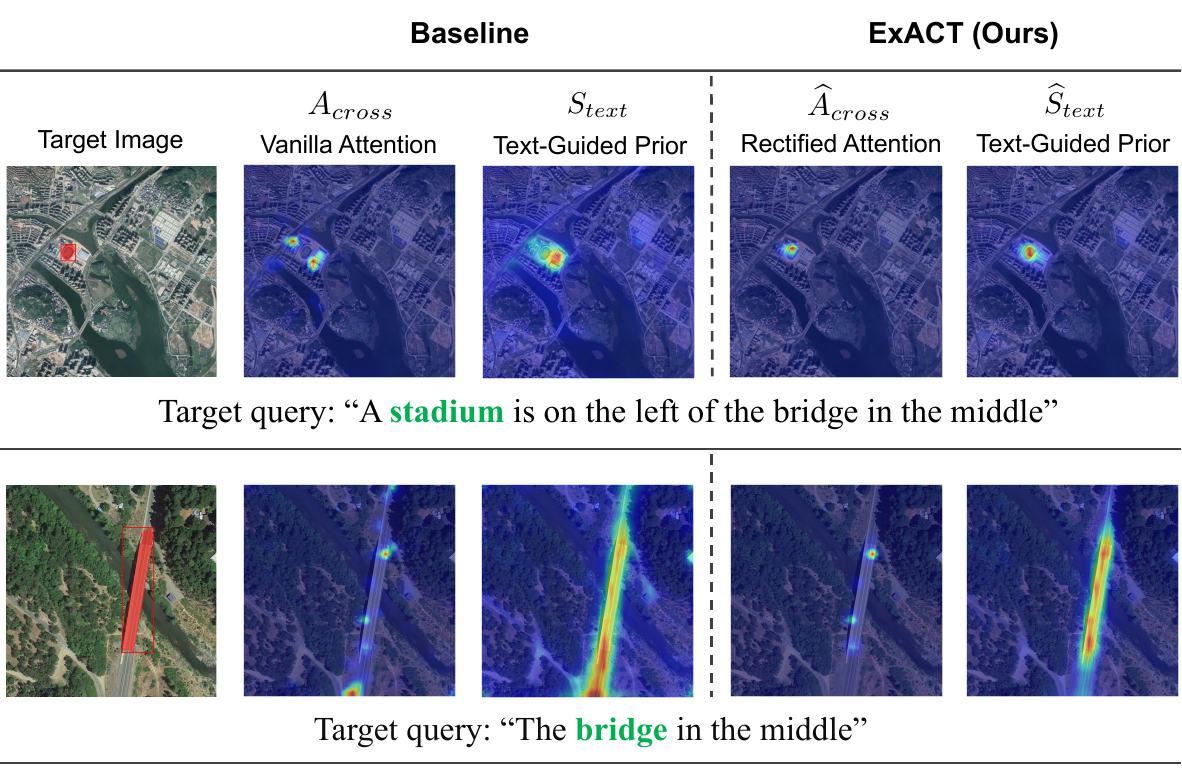}
  \caption{Effectiveness of our exemplar-driven calibration. The text-driven baseline suffers from spatial deviation ($A_{cross}$), which causes boundary leakage ($S_{text}$) after unconstrained diffusion. Our ExACT rectifies this misaligned prior, confining activations within the target region ($\widehat{A}_{cross}$, $\widehat{S}_{text}$).}
  \Description{The baseline suffers from noisy vanilla cross-attention ($A_{cross}$), leading to severe boundary leakage after unconstrained structural diffusion ($S_{text}$). Our ExACT calibrates the misaligned priors, ensuring that the activations are confined strictly to the target range ($\widehat{A}_{cross}$, $\widehat{S}_{text}$).}
  \label{text_prior}
\end{figure}

\begin{figure*}
  \centering
  \includegraphics[width=\linewidth]{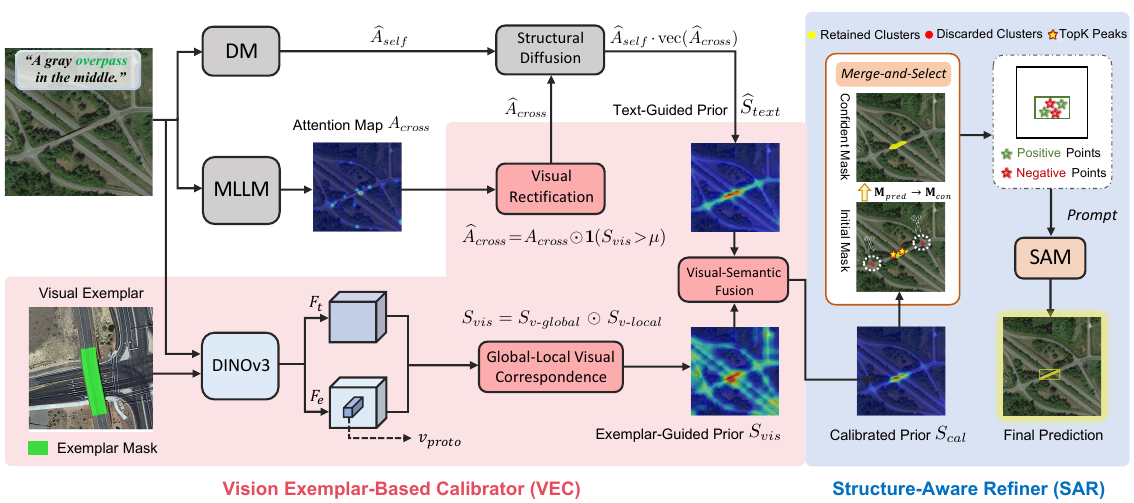}
  \caption{ExACT comprises two core modules: (1) The Vision Exemplar-Based Calibrator extracts dense feature-level correlations from a one-shot visual exemplar to rectify the noisy cross-modal priors $A_{cross}$, yielding a calibrated boundary-constrained localization map $S_{cal}$. (2) The Structure-Aware Refiner distills $S_{cal}$ via a \textit{merge-and-select} pipeline into high-quality geometric prompts that guide SAM for precise pixel-level grounding.}
  \Description{The GLVC mechanism extracts dense cross-image visual correspondences via DINOv3, yielding the exemplar-guided prior $S_{vis}$. This prior filters the coarse MLLM attention $A_{cross}$ via VR, producing a clean $\widehat{A}_{cross}$ that guides structural diffusion to recover local details. Through the VSF mechanism, the resulting text-guided prior $\widehat{S}_{text}$ fuses with $S_{vis}$ to form the calibrated unified prior $S_{cal}$. Finally, SAR consolidates $S_{cal}$ via a \textit{merge-and-select} pipeline into high-quality geometric prompts that guide SAM for precise pixel-level grounding.}
  \label{framework}
\end{figure*}

\section{ExACT Framework}

\subsection{Training-Free RSVG}
\label{sec:preliminaries}
Given an RS image $\mathcal{I}$ and a textual query $\mathcal{Q}$, standard training-free RSVG aims to localize the target without parameter updates, generating a pixel-level segmentation mask $\mathbf{M}_{pred} \in \{0, 1\}^{H \times W}$. This paradigm, which extracts coarse localization priors from frozen MLLMs and recovers structural details via off-the-shelf DMs, serves as the foundation of our work.

\textbf{(1) Language-Guided Coarse Localization.} To extract spatial cues, frozen MLLMs are prompted to formulate localization as a coordinate sequence generation task. During the autoregressive generation, the sentence-level cross-modal prior $A_{cross} \in \mathbb{R}^{h \times w}$ is extracted by averaging the visual attention weights across all attention heads, generation steps, and localization layers. Although $A_{cross}$ highlights the approximate target location, it inherently exhibits coordinate-driven bias and high-activation artifacts, as shown in the $A_{cross}$ column of Figure. \ref{text_prior}.

\textbf{(2) Diffusion-Based Structural Recovery.} To refine this coarse prior, the baseline leverages dense pixel affinities encoded in the self-attention maps from a frozen UNet-based DM. A unified structural prior $\widehat{A}_{self} \in \mathbb{R}^{hw \times hw}$ is obtained by averaging self-attention maps across all layers. The final text-guided prior $S_{text}$ is then computed by diffusing the coarse cues $A_{cross}$ across the entire structure:
\begin{equation}
    S_{text} = \text{Norm}\left( \widehat{A}_{self} \cdot \text{vec}(A_{cross}) \right) \in \mathbb{R}^{h \times w},
\end{equation}
where $\text{vec}(\cdot)$ denotes vectorization, $\text{Norm}(\cdot)$ applies Min-Max normalization, and the output is reshaped to  $h \times w$.

As observed in Figure \ref{text_prior}, applying such unconstrained structural completion directly to the noisy vanilla attention $A_{cross}$ inevitably amplifies background artifacts, triggering severe boundary leakage in $S_{text}$. Conversely, our proposed ExACT rectifies this misaligned prior using a one-shot visual exemplar. This rectification ($\widehat{A}_{cross}$) confines structural propagation strictly within the target region, yielding highly precise activations without spatial spillover, as shown by $\widehat{S}_{text}$.

\subsection{Framework Overview}
\label{sec:overview}
To address the spatio-semantic misalignment inherent in purely text-driven paradigms, our proposed ExACT framework introduces a one-shot visual exemplar, comprising an exemplar image and its corresponding mask, to provide explicit structural guidance. 
Figure \ref{framework} illustrates the overall workflow of ExACT. Following \cite{rsvg-zero}, we first extract a coarse cross-modal prior $A_{cross}$ from an off-the-shelf MLLM and a diffusion structural prior $\widehat{A}_{self}$ from a frozen DM. Importantly, we adopt a two-fold pipeline to achieve precise pixel-level grounding, focusing on localization calibration and mask refinement.
To calibrate spatial localization, the Vision Exemplar-Based Calibrator (Sec. \ref{sec:vec}) operates through three synergistic mechanisms. First, Global-Local Visual Correspondence (GLVC) leverages the frozen DINOv3 encoder to extract dense feature-level correlations between the visual exemplar and the target image. This produces a precise exemplar-guided prior $S_{vis}$ that explicitly highlights the structural extent of the target category. Second, through Visual Rectification (VR), $S_{vis}$ acts as an explicit structural gate to filter the noisy $A_{cross}$ into a clean $\widehat{A}_{cross}$. This rectified attention then guides structural diffusion alongside $\widehat{A}_{self}$ to recover local details, yielding a refined text-guided prior $\widehat{S}_{text}$. Finally, a Visual-Semantic Fusion (VSF) mechanism combines these two complementary modalities ($\widehat{S}_{text}$ and $S_{vis}$) into a boundary-constrained localization map $S_{cal}$. For mask refinement, the Structure-Aware Refiner (Sec. \ref{sec:sar}) consolidates fragmented regions and suppresses spurious activations, extracting high-quality geometric prompts for precise SAM-driven pixel-level localization.

\subsection{Vision Exemplar-Based Calibrator}
\label{sec:vec}

Linguistic cues inherently struggle to convey fine-grained visual details necessary to delineate precise object boundaries and distinguish subtle appearance differences in cluttered RS scenes. To bridge this modality gap, accurate localization requires explicit visual features to serve as definitive references. Accordingly, our Vision Exemplar-Based Calibrator (VEC) is designed to supply discriminative structural guidance, effectively resolving spatial ambiguity and geometric uncertainty. As detailed below, this module operates through three core mechanisms: Global-Local Visual Correspondence (GLVC), Visual Rectification (VR), and Visual-Semantic Fusion (VSF).

\noindent\textbf{Global-Local Visual Correspondence.}
Recognizing that global-average matching overlooks intra-class variations while local matching is vulnerable to background clutter, GLVC effectively combines global structural consistency with local detail verification.

Given a readily accessible one-shot visual exemplar $(I_e, M_e)$ for each category, we employ the pre-trained DINOv3 image encoder to extract dense patch-level features $F_t, F_e \in \mathbb{R}^{h \times w \times C}$ from the target and exemplar images.
To prevent losing small objects during standard downsampling, we introduce a \textit{foreground-preserving masking} strategy to construct an exemplar foreground feature set $\mathcal{F}_e$.

To capture the global structural consensus of the target category, we compute a global visual prototype $v_{proto}$ by average-pooling the features in $\mathcal{F}_e$. The global correlation map $S_{v\text{-}global} \in \mathbb{R}^{h \times w}$ is then obtained via cosine similarity between this prototype vector $v_{proto}$ and each target feature $f_t^{(i,j)} \in F_t$:

\begin{equation}
    S_{v\text{-}global}^{(i,j)} = \frac{f_t^{(i,j)} \cdot v_{proto}}{\|f_t^{(i,j)}\|_2 \|v_{proto}\|_2}.
\end{equation}

To verify fine-grained details (e.g., edges and parts), we also perform dense patch-level retrieval. By assigning each target feature its maximum similarity within the exemplar foreground feature set $\mathcal{F}_e$, we produce the local correlation map $S_{v\text{-}local} \in \mathbb{R}^{h \times w}$:
\begin{equation}
    S_{v\text{-}local}^{(i,j)} = \max_{f \in \mathcal{F}_e} \left( \frac{f_t^{(i,j)} \cdot f}{\|f_t^{(i,j)}\|_2 \|f\|_2} \right).
\end{equation}

Finally, to harness their complementarity, we obtain a comprehensive exemplar-guided prior $S_{vis} \in \mathbb{R}^{h \times w}$ through element-wise multiplication:
\begin{equation}
     S_{vis} = \text{Norm}(S_{v\text{-}global} \odot S_{v\text{-}local}),
\end{equation}
where $\odot$ denotes the Hadamard product and $\text{Norm}(\cdot)$ performs Min-Max normalization. This geometric intersection suppresses background clutter while reinforcing the consistency of structural boundaries, serving as a highly discriminative visual constraint.

\begin{figure*}[!h]
    \centering
    \includegraphics[width=\linewidth]{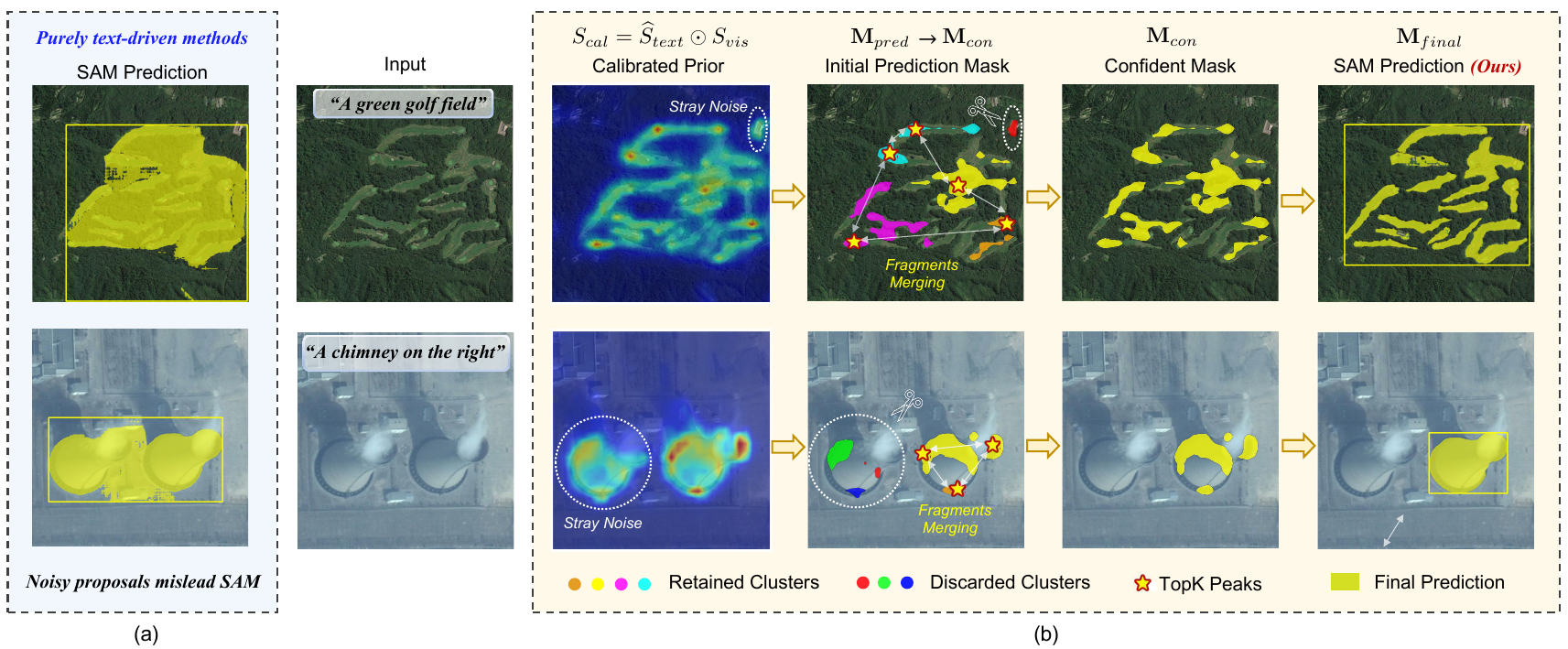}
  \vspace{-0.3cm}
	\caption{Illustration of Structure-Aware Refiner. (a) Text-driven baselines yield noisy proposals that mislead SAM. (b) SAR resolves this via a \textit{merge-and-select} mechanism: adaptive-scale geometric aggregation first reconnects physical target fragments into cohesive clusters (distinct colors). Semantic selection then retains valid clusters with Top-K semantic anchors (stars, linked by white arrows) while discarding isolated noise (scissors), yielding a confident mask ($\mathbf{M}_{con}$) for precise SAM grounding.}
  \Description{Illustration of Structure-Aware Refiner. (a) Text-driven baselines yield noisy proposals that mislead SAM. (b) SAR resolves this via a \textit{merge-and-select} mechanism: adaptive-scale geometric aggregation first reconnects physical target fragments into cohesive clusters (distinct colors). Semantic selection then retains valid clusters with Top-K semantic anchors (stars, linked by white arrows) while discarding isolated noise (scissors), yielding a confident mask ($\mathbf{M}_{con}$) for precise SAM grounding.}
\label{SAR}
\end{figure*}

\noindent\textbf{Visual Rectification.} The vanilla MLLM prior $A_{cross}$ often suffers from coordinate-driven bias and background artifacts. To correct this, VR employs the exemplar-guided prior $S_{vis}$ as a deterministic structural gate to prune noisy text-driven activations. 

Formally, the rectified attention map $\widehat{A}_{cross} \in \mathbb{R}^{h \times w}$ is computed via adaptive hard-gating:
\begin{equation}
    \widehat{A}_{cross} = A_{cross} \odot \mathds{1}(S_{vis} > \mu),
\end{equation}
where $\mathds{1}(\cdot)$ denotes the indicator function, and $\mu$ represents the spatial mean of $S_{vis}$.

This gating mechanism suppresses dominant anomalies, yielding a clean $\widehat{A}_{cross}$ that redirects attention from sparse extremities to the true structural range of the queried objects. To further compensate for incomplete activations, we apply structural diffusion on $\widehat{A}_{cross}$. This produces an improved text-guided prior $\widehat{S}_{text}$ with coherent boundaries, preparing for the final visual-semantic fusion.

\noindent\textbf{Visual-Semantic Fusion} 
 Although the text-guided prior $\widehat{S}_{text}$ successfully recovers structural details, DMs inherently prioritize low-level spatial continuity. 
This often leads to boundary leakage, where generative activations erroneously diffuse into surrounding contexts (e.g., leaking from the ``overpass'' onto connecting roads, as seen in Fig. \ref{framework}). To mitigate this, VSF leverages the discriminative exemplar-guided prior $S_{vis}$ to verify and constrain these generative activations. Formally, the final calibrated prior $S_{cal} \in \mathbb{R}^{h \times w}$ is computed through cross-modal structural verification:
\begin{equation}
    S_{cal} = \widehat{S}_{text} \odot S_{vis}.
\end{equation}
This Hadamard product acts as a strict spatial pruner, removing anomalous regions that lack clear visual support. The resulting $S_{cal}$ is a semantically consistent, boundary-constrained localization map, providing a high-confidence geometric foundation for subsequent mask refinement.

\subsection{Structure-Aware Refiner}
\label{sec:sar}

Binarizing the calibrated prior $S_{cal}$ with a threshold $\alpha$ yields an initial mask $\mathbf{M}_{pred}$. However, driven by scattered attention patterns, $\mathbf{M}_{pred}$ often suffers from physical fragmentation and outlier noise. Thus, utilizing this raw mask directly to prompt SAM is suboptimal, resulting in noisy proposals that could mislead SAM into segmenting outliers, as illustrated in Figure. \ref{SAR}(a). To extract high-quality geometric prompts for SAM, our SAR module employs a coarse-to-fine \textbf{\textit{merge-and-select}} pipeline.

\noindent\textbf{Iterative Geometric Aggregation}
Since fixed-distance merging fails across varying RS scales, we perform adaptive-scale clustering to reconnect physical fragments. We first decompose $\mathbf{M}_{pred}$ into disjoint connected components through morphological operations. Starting with the largest component $\hat{c}$, we compute a dynamic spatial radius $\delta_{dyn} = \frac{1}{|\hat{c}|} \sum_{p \in \hat{c}} \|p - \bar{p}\|_2$, where $\bar{p}$ is its centroid. The cluster iteratively absorbs neighboring fragments within $\delta_{dyn}$, grouping all fragments into cohesive candidate clusters $\mathcal{G} = \{G_1, \dots, G_m\}$.  As shown in Figure. \ref{SAR} ($\mathbf{M}_{pred} \rightarrow \mathbf{M}_{con}$), distinct colors denote individual clusters, effectively consolidating fragmented regions within the geometric range of the instance.

\noindent\textbf{TopK-Guided Semantic Selection} 
To filter out ambiguous distractors from multiple independent geometric clusters $\mathcal{G}$, we extract high-confidence semantic anchors $\mathcal{P}_{topk} = \operatorname{TopK}(S_{cal}, K)$ from the calibrated prior.
We retain only the clusters containing at least one Top-$K$ peak, while strictly discarding those without semantic support. The final confident mask $\mathbf{M}_{con}$ is formed by the union of these selected clusters:
\begin{equation}
    \mathbf{M}_{con} = \bigcup_{G_i \in \mathcal{G}} \{ G_i \mid G_i \cap \mathcal{P}_{topk} \neq \emptyset \}.
\end{equation}
This selection mechanism ensures that $\mathbf{M}_{con}$ preserves the complete semantic body while eliminating stray outlier noise.

Finally, the consolidated mask $\mathbf{M}_{con}$ is utilized to extract a bounding box along with representative positive and negative point prompts. These geometric prompts robustly guide SAM to yield the precise pixel-level prediction.

\begin{table*}
  \caption{Comparison with the latest SOTA methods on the RRSIS-D dataset. ``-'' indicates not using MLLMs.}
  \label{tab:rrsis-d}
  \vspace{-0.3cm}
  \renewcommand{\arraystretch}{1}
  \setlength{\tabcolsep}{2.5pt}
  \small
  \resizebox{\textwidth}{!}{
  \begin{tabular}{lcccccccccccc}
    \toprule
    \multirow{2}{*}{Method} & \multirow{2}{*}{\makecell{Pretrained\\MLLMs}} & \multirow{2}{*}{\makecell{Training-\\Free}}   & \multicolumn{5}{c}{RSREC} & \multicolumn{5}{c}{RSRES} \\ 
\cmidrule(r){4-8} \cmidrule(r){9-13}
     & & & PR@0.3 & PR@0.5 & PR@0.7 & mIoU & oIoU & PR@0.3 & PR@0.5 & PR@0.7 & mIoU & oIoU \\ 
\midrule 
\rowcolor{titlegray}
\multicolumn{13}{l}{\textbf{Weakly-supervised text-driven models}}\\
    TRIS\cite{TRIS}\textcolor{venuegray}{\textit{\scriptsize ICCV'23}} &- & \ding{55} & 14.62 & 3.79 & 0.37 & 13.20 & 13.14 & 15.02 & 4.91 & 1.26 & 13.11 & 15.40 \\ 
    SAG\cite{SAG}\textcolor{venuegray}{\textit{\scriptsize ICCV'23}} &- &\ding{55} & 9.39 & 2.10 & 0.29 & 9.22 & 9.21 & 11.63 & 4.02 & 0.60 & 11.10 & 11.13 \\ 
    QueryMatch\cite{querymatch}\textcolor{venuegray}{\textit{\scriptsize ACMMM'24}} &- & \ding{55} & 22.04 & 16.22 & 12.10 & 17.21 & 15.26 & 20.97 & 15.54 & 10.62 & 15.73 & 10.73 \\ 
\midrule
\rowcolor{titlegray}
\multicolumn{13}{l}{\textbf{Zero-shot text-driven models}}\\
    LISA\cite{lisa}\textcolor{venuegray}{\textit{\scriptsize CVPR'24}} & LLaVA2-13B & \ding{55} & 36.23 & 22.52 & 12.47 &  26.72 & 30.01 &  41.05 & 27.41 & 16.03 & 29.98 &  35.18\\ 
    GroundVLP\cite{groundvlp}\textcolor{venuegray}{\textit{\scriptsize AAAI'24}}  &- & \ding{55} & 21.26 & 16.20 & 11.32 & 16.51 & 16.88 & 18.07 & 13.33 & 7.24 & 13.14 &11.23\\
    GeoChat\cite{geochat}\textcolor{venuegray}{\textit{\scriptsize CVPR'24}}  + SAM &Vicuna1.5-7B & \ding{55} & 39.51 & 26.94 &  12.05& 30.46 &  28.58 & 34.25 & 18.84 & 10.61 & 24.07 & 19.32  \\ 
    Qwen2.5-VL\cite{qwen2}\textcolor{venuegray}{\textit{\scriptsize arXiv'24}} + SAM  & Qwen2.5-VL-7B & \ding{55} & 40.30 & 30.54 & 15.67 & 32.15 & 29.14 & 35.88 & 24.52 & 12.57 & 26.22 &  20.51\\
    RSVG-ZeroOV\cite{rsvg-zero} \textcolor{venuegray}{\textit{\scriptsize AAAI'26}}  & Qwen2.5-VL-7B & \ding{51} & 45.70 & 31.39 & 17.63 & 34.49 & 31.28 & 40.01 & 27.39 &
     13.38 &  28.35 & 22.83  \\
\midrule 
\rowcolor{titlegray}
\multicolumn{13}{l}{\textbf{One-shot exemplar-driven  models}}\\
    RSVG-ZeroOV (\emph{w/ VSF}) & Qwen2.5-VL-7B & \ding{51} & 50.04 & 38.44 & 28.58  & 39.06 & 34.87 & 49.55 & 38.29 & 26.20 & 37.24 & 29.66 \\
\rowcolor{ourspink}
Ours & Qwen2.5-VL-7B  & \ding{51} & \textbf{52.31} & \textbf{39.82} & \textbf{28.58} & \textbf{40.19} & \textbf{40.79} & \textbf{52.71} & \textbf{39.82} & \textbf{26.98} & \textbf{39.29} & \textbf{38.52} \\
\bottomrule
  \end{tabular}}
\end{table*}

\begin{table*}
  \caption{Comparison with the latest SOTA methods on the RISBench dataset. ``-'' indicates not using MLLMs.}
  \label{tab:risbench}
  \vspace{-0.3cm}
  \renewcommand{\arraystretch}{1}
  \setlength{\tabcolsep}{2.5pt}
  \small
  \resizebox{\textwidth}{!}{
  \begin{tabular}{lcccccccccccc}
    \toprule
    \multirow{2}{*}{Method} & \multirow{2}{*}{\makecell{Pretrained\\MLLMs}} & \multirow{2}{*}{\makecell{Training-\\Free}}   & \multicolumn{5}{c}{RSREC} & \multicolumn{5}{c}{RSRES} \\ 
\cmidrule(r){4-8} \cmidrule(r){9-13}
     & & & PR@0.3 & PR@0.5 & PR@0.7 & mIoU & oIoU & PR@0.3 & PR@0.5 & PR@0.7 & mIoU & oIoU \\ 
\midrule 
\rowcolor{titlegray}
\multicolumn{13}{l}{\textbf{Weakly-supervised text-driven models}}\\
    TRIS\cite{TRIS}\textcolor{venuegray}{\textit{\scriptsize ICCV'23}} &- & \ding{55} & 19.27 & 8.89 & 2.86 & 14.63 & 15.74 & 14.23 & 4.40 & 0.92 & 11.46 & 16.06 \\ 
    SAG\cite{SAG}\textcolor{venuegray}{\textit{\scriptsize ICCV'23}} &- &\ding{55} & 5.83 & 1.78 & 0.33 & 7.12 & 7.10 & 8.79 & 2.55 & 0.35 & 9.31 & 9.36 \\ 
    QueryMatch\cite{querymatch}\textcolor{venuegray}{\textit{\scriptsize ACMMM'24}} &- & \ding{55} & 31.79 & 27.74 & 23.32 & 26.72 & 15.86 & 31.06 & 26.68 & 20.27 & 24.59 & 10.77 \\ 
\midrule
\rowcolor{titlegray}
\multicolumn{13}{l}{\textbf{Zero-shot text-driven models}}\\
    LISA\cite{lisa}\textcolor{venuegray}{\textit{\scriptsize CVPR'24}} & LLaVA2-13B & \ding{55} &34.84 & 24.03 &16.54 &28.38 & 31.12 & 39.88 & 27.55 & 17.43 & 30.05 & 31.64 \\ 
    GroundVLP\cite{groundvlp}\textcolor{venuegray}{\textit{\scriptsize AAAI'24}}  &- & \ding{55} & 23.13 & 19.91 & 16.37 & 19.19 & 16.17 & 20.78 & 15.82 & 10.95 & 15.58 &10.08\\
    GeoChat\cite{geochat}\textcolor{venuegray}{\textit{\scriptsize CVPR'24}}  + SAM &Vicuna1.5-7B & \ding{55} &45.88 &34.85 &16.89 &36.57 &29.55 &37.25 &25.94 &14.17 &28.22 &21.49  \\
    Qwen2.5-VL\cite{qwen2}\textcolor{venuegray}{\textit{\scriptsize arXiv'24}} + SAM  & Qwen2.5-VL-7B &\ding{55} &47.28 &37.95 &22.36 &37.82 &30.84 &40.68 &30.38 &15.33 &30.67 &20.98 \\
    RSVG-ZeroOV\cite{rsvg-zero} \textcolor{venuegray}{\textit{\scriptsize AAAI'26}}  & Qwen2.5-VL-7B & \ding{51}  &50.77 & 38.90 & 24.93 & 38.87 & 34.30 & 44.30 & 31.03 & 18.61 &  31.84 & 26.35  \\ 
\midrule 
\rowcolor{titlegray}
\multicolumn{13}{l}{\textbf{One-shot exemplar-driven  models}}\\
    RSVG-ZeroOV (\emph{w/ VSF}) & Qwen2.5-VL-7B & \ding{51} &  53.94 & 40.73 & 30.48 & 41.28 & 38.26 & 52.34 & 39.72 & 27.58 & 40.21 & 31.86  \\
\rowcolor{ourspink}
Ours & Qwen2.5-VL-7B  & \ding{51}  & \textbf{56.41} & \textbf{45.34} & \textbf{33.10} & \textbf{44.17} & \textbf{43.54} & \textbf{55.69} & \textbf{45.68} & \textbf{31.41} & \textbf{42.43} &  \textbf{41.58}  \\
\bottomrule
  \end{tabular}}
\end{table*}

\section{Experiments} 
\subsection{Experimental Settings}
\noindent\textbf{Datasets.} 
We evaluate the performance of ExACT on RRSIS-D \cite{RMSIN} and RISBench \cite{CroBIM} benchmarks, which provide comprehensive annotations to support both RSREC and RSRES tasks.
Further details are provided in the Appendix. 

\noindent\textbf{Implementation Details.}
Our proposed ExACT functions as a training-free, one-shot visual prompting solution that requires only forward inference, without parameter optimization. For patch-level feature extraction, we utilize the frozen DINOv3 \cite{dino} with a ViT-B/16 backbone. Consistent with the baseline \cite{rsvg-zero}, we employ the pre-trained Qwen2.5-VL \cite{qwen2} to generate the coarse cross-modal prior, and Stable Diffusion V1.4 \cite{stable-diffusion} to extract structural priors. SAM \cite{sam} with a ViT-H backbone serves as our off-the-shelf segmenter. Additionally, we set the binarization threshold to $\alpha = 0.4$ and the number of semantic anchors to $K = 15$. Detailed configurations for geometric prompt generation are provided in the Appendix. All experiments are conducted on a single NVIDIA RTX 3090 GPU.

\subsection{Comparison with State-of-the-Arts}
\label{sec:comparison}
Tables \ref{tab:rrsis-d} and \ref{tab:risbench} present the quantitative comparison between our ExACT and state-of-the-art methods on the RRSIS-D and RISBench benchmarks. We categorize the compared methods into three distinct paradigms: (1) weakly-supervised text-driven methods (e.g., QueryMatch \cite{querymatch}), (2) zero-shot text-driven models (e.g., LISA \cite{lisa}, RSVG-ZeroOV \cite{rsvg-zero}), and (3) one-shot exemplar-driven approaches (e.g., RSVG-ZeroOV \textit{w/ VSF}, an exemplar-augmented baseline constructed for a fair comparison).

\noindent\textbf{Comparison on RRSIS-D.}
As shown in Table \ref{tab:rrsis-d}, ExACT achieves SOTA performance on both RSREC and RSRES tasks. Compared to the strongest text-driven baseline RSVG-ZeroOV, our approach delivers substantial IoU improvements of +9.51\% and +15.69\% on RSREC and RSRES, respectively. This highlights the inherent spatial ambiguity of purely text-guided grounding and validates our exemplar-driven calibration.
Furthermore, while equipping this baseline with explicit visual guidance (RSVG-ZeroOV \textit{w/ VSF}) yields notable mIoU gains (+4.57\% and +10.94\%), ExACT consistently outperforms it. Notably, ExACT achieves 39.29\% mIoU and 38.52\% oIoU on RSRES, narrowing the gap between them to 0.77\%. This also confirms the effectiveness of our method in suppressing false positives and achieving precise pixel-level grounding.

\noindent\textbf{Comparison on RISBench. }
As presented in Table \ref{tab:risbench}, ExACT continues to outperform all others across all metrics on the more challenging RISBench dataset. Traditional models like QueryMatch and GroundVLP fall behind significantly, with 17.84\% and 26.85\% drops in RSRES mIoU, respectively, due to their inability to capture detailed object structures. Against the training-free text-driven RSVG-ZeroOV, ExACT achieves a massive +12.80\% margin on the strict PR@0.7 metric for RSRES (31.41\% vs. 18.61\%).
This contrast exposes the limitations of purely text-guided priors for precise boundary delineation. While adding visual guidance to the baseline (RSVG-ZeroOV \textit{w/ VSF}) improves mIoU by +2.41\% and +8.37\% on RSREC and RSRES, respectively, our full framework increases the overall RSRES IoU to 41.58\% (an additional +9.72\% gain over the augmented baseline). This underscores the critical role of our structure-aware geometric prompts in facilitating coarse-to-fine mask refinement.

\begin{table}[htbp]
    \centering
    
    \caption{Ablation study on core components.}
    \label{tab:components}
    \vspace{-0.3cm}
    \resizebox{0.34\textwidth}{!}{
    \begin{tabular}{ccccc}
        \toprule
        \multirow{2.5}{*}{Variant} & \multicolumn{2}{c}{RSREC} & \multicolumn{2}{c}{RSRES} \\
        \cmidrule(lr){2-3} \cmidrule(lr){4-5}
        & Pr@0.5 & mIoU & Pr@0.5 & mIoU \\
        \midrule
        w/o VR & 37.66 & 38.28 & 38.12 & 37.56 \\
        w/o VSF & 33.75 & 35.79 & 34.56 & 35.24 \\
        w/o SAR & 33.32 & 34.98 & 33.98 & 33.96 \\
        \rowcolor{ourspink}
        Full Model & \textbf{39.82} & \textbf{40.19} & \textbf{39.82} & \textbf{39.29} \\
        \bottomrule
    \end{tabular}}

    \vspace{0.25cm} 

    \caption{Ablation study on visual exemplar quantity ($n$-shot).}
    \label{tab:n_shot}
    \vspace{-0.3cm}
    \begin{tabular}{ccccc}
        \toprule
        \multirow{2.5}{*}{$n$-shot} & \multicolumn{2}{c}{RSREC} & \multicolumn{2}{c}{RSRES} \\
        \cmidrule(lr){2-3} \cmidrule(lr){4-5}
        & Pr@0.5 & mIoU & Pr@0.5 & mIoU \\
        \midrule
        \rowcolor{ourspink}
       1-shot & \textbf{39.82} & 40.19 & 39.82 & 39.29 \\
        3-shot & 39.27 & 39.94 & 39.41 & 39.46 \\
        5-shot & 39.44 & \textbf{40.34} & \textbf{39.90} & \textbf{39.64}\\
        \bottomrule
    \end{tabular}






\end{table}

\subsection{Ablation Studies}
\noindent\textbf{Ablation of Core Components.} Table \ref{tab:components} confirms the effectiveness of the core modules in ExACT on the RRSIS-D dataset by evaluating three separate variants: (i) w/o VR, which depends directly on noisy cross-attention; (ii) w/o VSF, which bypasses visual-semantic fusion; and (iii) w/o SAR, which replaces the \textit{merge-and-select} refinement with standard thresholding. Removing VR reduces the RSREC mIoU from 40.19\% to 38.28\%, revealing that uncalibrated MLLM priors are affected by coordinate-driven bias and high-activation artifacts. Disabling VSF results in a steeper decrease, degrading Pr@0.5 by 5\%-6\%. This demonstrates that heavy reliance on text-guided generative structural cues can lead to boundary leakage, underscoring the importance of explicit visual verification. Notably, the w/o SAR variant performs the worst, dropping to 34.98\% and 33.96\% mIoU for RSREC and RSRES, respectively. This severe decline highlights that transitioning from continuous maps to discrete prompts is a major structural bottleneck. Ultimately, our complete ExACT model addresses this via structure-aware refinement, effectively filtering out stray noise to achieve precise grounding.

\noindent\textbf{Ablation of Visual Exemplar Quantity.} Table \ref{tab:n_shot} evaluates how the number of visual exemplars ($n$-shot) affects performance on the RRSIS-D dataset. While 5-shot inference provides slight mIoU improvements (+0.15\% and +0.35\% on RSREC and RSRES), the 1-shot setup remains highly competitive, reaching a peak Pr@0.5 of 39.82\% Pr@0.5 on RSREC. To prevent redundant feature matching and unnecessary computational costs, ExACT adopts the 1-shot strategy as its default setting, ensuring an optimal trade-off between accuracy and efficiency.

\begin{figure}[!h]
	\centering
	\includegraphics[scale=0.63]{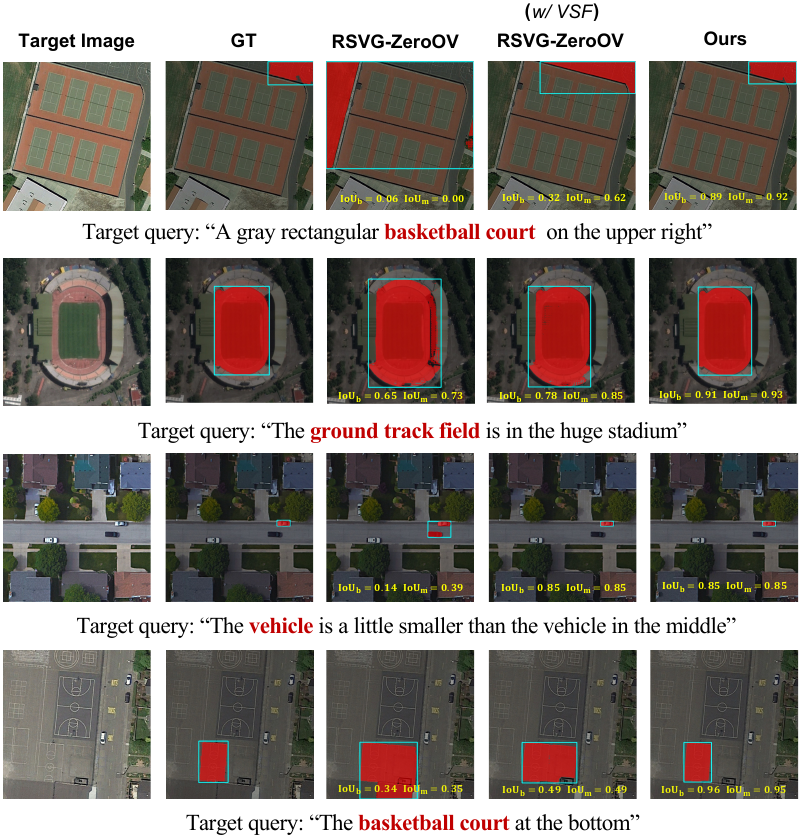}
  \vspace{-0.5cm}
	\caption{Qualitative results on the RRSIS-D dataset. The prediction bounding box $\text{IoU}_{\text{b}}$ and mask $\text{IoU}_{\text{m}}$  scores are indicated in yellow.}
   \Description{Qualitative results on the RRSIS-D dataset. The prediction bounding box $\text{IoU}_{\text{b}}$ and mask $\text{IoU}_{\text{m}}$  scores are indicated in yellow.}
	\label{results}
\end{figure}

\begin{figure*}[!h]
	\includegraphics[scale=0.65]{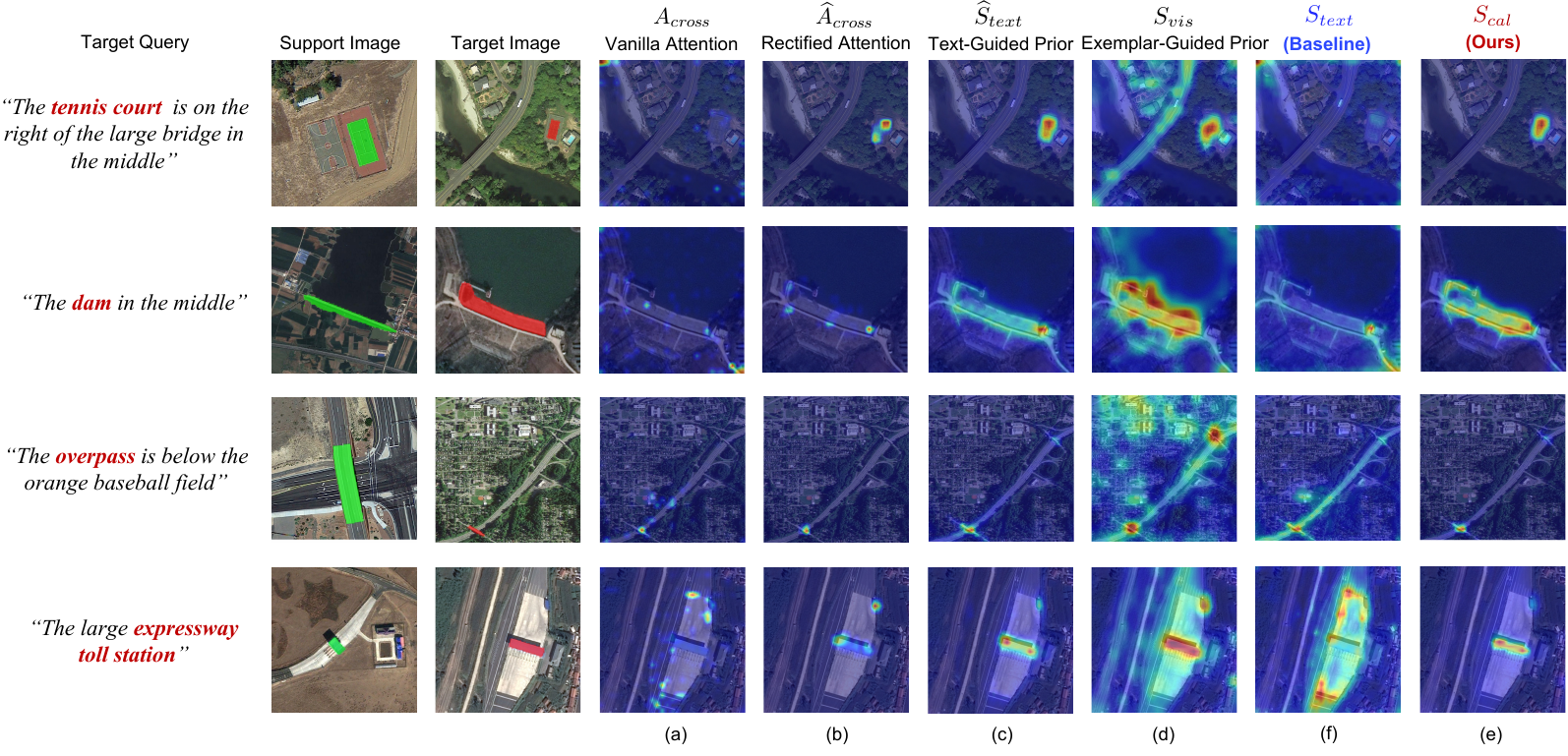}
  \vspace{-0.3cm}
	\caption{Step-by-step visualization of the internal localization priors. ExACT progressively rectifies noisy attention (a-b) and fuses visual-semantic cues (c-d) to generate a highly focused calibrated prior (f), outperforming the diffuse baseline (e).}
  \Description{(a)-(b) Initial MLLM attention $A_{cross}$ and the rectified activation $\widehat{A}_{cross}$ after VR. (c) Text-guided prior $\widehat{S}_{text}$. (d) Vision-guided prior $S_{vis}$ (from reference exemplars). (e)-(f) Comparison of final response maps between the baseline $S_{text}$ (RSVG-ZeroOV\cite{rsvg-zero}) and our $S_{cal}$. While MLLM attention maps provide a rough localization prior with limited precision, the vision-guided prior captures fine-grained semantic cues but risks highlighting similar distractors. By leveraging VR and VSF mechanisms to integrate these complementary priors, our method achieves accurate and semantically consistent localization.}
	\label{heatmaps}
\end{figure*}

\begin{table}[!h]
    \centering
    \vspace{0.25cm} 

    \caption{Ablation study on MLLM backbones.}
    \label{tab:MLLM}
    \vspace{-0.3cm}
    \begin{tabular}{ccccc}
        \toprule
        \multirow{2.5}{*}{MLLM} & \multicolumn{2}{c}{RSREC} & \multicolumn{2}{c}{RSRES} \\
        \cmidrule(lr){2-3} \cmidrule(lr){4-5}
        & Pr@0.5 & mIoU & Pr@0.5 & mIoU \\
        \midrule
       LLaVA1.5-13B & 35.02 & 36.11 & 34.85 & 34.79 \\
       \rowcolor{ourspink}
        Qwen2.5-VL-7B &\textbf{39.82} & \textbf{40.19} & \textbf{39.82} & \textbf{39.29} \\

        \bottomrule
    \end{tabular}

    \vspace{0.25cm} 

    \caption{Ablation study on global-local matching.}
    \label{global-local}
    \vspace{-0.3cm}
    \begin{tabular}{cccccc}
        \toprule
       \multirow{2.5}{*}{Global} & \multirow{2.5}{*}{Local}    & \multicolumn{2}{c}{RSREC} & \multicolumn{2}{c}{RSRES} \\
        \cmidrule(lr){3-4} \cmidrule(lr){5-6}
       & & Pr@0.5 & mIoU & Pr@0.5 & mIoU \\
        \midrule
        \ding{51} & & 37.14 & 38.24 & 37.46 & 37.63 \\
         & \ding{51} & 36.71 & 38.07 & 36.94 & 37.13 \\
      \rowcolor{ourspink}
      \ding{51} & \ding{51} & \textbf{39.82} & \textbf{40.19} & \textbf{39.82} & \textbf{39.29} \\
        \bottomrule
    \end{tabular}

\end{table}

\noindent\textbf{Ablation of MLLM Backbones.}
Table \ref{tab:MLLM} assesses the generalizability of ExACT across different MLLM architectures. Notably, switching the backbone from LLaVA1.5-13B to Qwen2.5-VL-7B results in significant improvements in mIoU of +4.08\% for RSREC and +4.50\% for RSRES. By performing well with a smaller model (7B vs. 13B), we demonstrate that our architecture-agnostic pipeline depends on intrinsic spatial alignment rather than raw parameter scale, enabling lightweight MLLMs to attain top performance.

\noindent\textbf{Ablation of Global-Local Matching}
Table \ref{global-local} examines the contributions of different visual matching granularities on the RRSIS-D dataset. Employing either global ($S_{v\text{-}global}$) or local  ($S_{v\text{-}lobal}$)  correspondence independently yields suboptimal results, achieving RSREC mIoU scores of 38.24\% and 38.07\%, respectively. Fusing both granularities into $S_{vis}$ drives a significant performance boost, reaching 40.19\% mIoU with gains of +1.95\% and +2.12\%. This improvement demonstrates that multi-scale correspondences are crucial for effective visual-semantic calibration in complex RS scenes.

\subsection{Qualitative Analysis}

\noindent\textbf{Qualitative Comparison.}
Figure \ref{results} presents representative visual comparisons on the RRSIS-D dataset.  Our approach consistently delivers the most accurate localization performance, while the VSF-augmented baseline remains competitive yet suboptimal. In scenes with multiple similar structures (e.g., rows 1 and 4), the baselines fail to distinguish the target from adjacent distractors, leading to severe false positives (e.g., misidentifying the wrong neighboring court). With the Vision Exemplar-Based Calibrator, ExACT filters out irrelevant regions lacking definitive visual support, refocusing the model's attention on the true target and boosting the mask IoU from 0.62 to 0.92 (row 1) and from 0.49 to 0.95 (row 4). Furthermore, for complex multi-object queries (rows 2 and 3), text-guided priors often yield coarse, over-segmented boundaries. Enabled by structure-aware refinement, ExACT effectively prevents instance adhesion and firmly encapsulates the target structures, reaching exceptional mask IoUs (e.g., 0.93 for the ground track field). Ultimately, these results confirm that by combining early-stage exemplar-driven calibration with structure-aware refinement, ExACT preserves the structural integrity and boundary accuracy of diverse RS targets in a training-free manner.

\noindent\textbf{Interpretability of Exemplar-Based Calibration.} To clarify the internal workings of ExACT, Figure. \ref{heatmaps} illustrates the step-by-step development of internal response maps. Initially, the raw MLLM cross-attention $A_{cross}$ (a) suffers from coordinate-driven bias and irrelevant background artifacts. Our VR mechanism explicitly rectifies this in $\widehat{A}_{cross}$ (b), shifting activations from corner extremities to the true structural footprint of the target. After structural diffusion, the text-guided prior $\widehat{S}_{text}$ (c) captures a rough object extent while effectively reducing background noise. Meanwhile, the vision-guided prior $S_{vis}$ (d) provides high-resolution details with sharper boundaries but lacks text-referring ability, which can sometimes cause multi-instance confusion (e.g., highlighting both overpasses in row 3). The Visual-Semantic Fusion elegantly resolves this conflict by using (c) as a semantic pruner to eliminate false positives in (d). As shown in the ``dam” and ``toll station” cases, our final fused map $S_{cal}$ (f) maintains a highly focused and physically accurate response, avoiding the diffuse and misaligned distribution seen in the baseline $S_{text}$ (e). This internal trajectory validates that the “rectification-then-fusion” pipeline in the Vision Exemplar-Based Calibrator effectively bridges text-guided localization and fine-grained visual perception.

\section{Conclusion}
This study proposes ExACT, a novel training-free framework that bridges the visual-semantic gap in purely text-driven RSVG by introducing a one-shot visual exemplar for discriminative structural guidance. ExACT utilizes a VEC module to explicitly rectify noisy MLLM priors and a SAR module to distill them into precise geometric structures. By synergizing MLLM semantic reasoning with VFM structural recovery, our method provides a robust, annotation-efficient pathway for open-vocabulary geospatial intelligence.  

\bibliographystyle{ACM-Reference-Format}
\bibliography{sample-base}

\clearpage
\appendix
\setcounter{table}{0}
\setcounter{figure}{0}
\twocolumn[
  \begin{center}
    {\LARGE\bfseries ExACT: Exemplar-Driven Calibrated Refinement for\\
    Training-Free Visual Grounding in Remote Sensing Images\par}
    \vspace{0.5em}
    {\large Supplementary Materials\par}
    \vspace{1.2em}
  \end{center}
]

\section{Geometric Prompt Generation}
\label{supp:prompt_gen}
The geometric prompt generation is mentioned in Sec.~3.4 of the main manuscript.
Here, we detail how high-quality box and point prompts are extracted from the
consolidated mask $\mathbf{M}_{con}$ to guide SAM toward the final accurate
prediction $\mathbf{M}_{final}$. Let $N$ and $\gamma$ denote the target number
of point prompts per instance and the oversampling factor, respectively.

\begin{itemize}
  \item \textbf{Box Prompt:} The bounding box $B_{box}$ is defined as the
  minimal axis-aligned rectangle enclosing the foreground region of
  $\mathbf{M}_{con}$.
  \item \textbf{Positive Points Prompt:} We first sample the $\gamma N$ pixels
  with the highest confidence from $\mathbf{M}_{con}$ and then perform K-Means
  clustering on these candidates to obtain $N$ spatial centroids. These
  centroids act as strong positive prompts.
  \item \textbf{Negative Points Prompt:} To prioritize hard negatives, we
  identify unique background pixels within
  $B_{box}\setminus\mathbf{M}_{con}$ with scores below a threshold
  $\tau_{neg}$. From this candidate pool, the $\gamma N$ points spatially
  nearest to $\mathbf{M}_{con}$ are selected, and the $N$ samples with the
  lowest activation scores are finally retained as negative prompts.
\end{itemize}
In summary, this strategy distills $\mathbf{M}_{con}$ into robust geometric
guidance for SAM. The combination of global box bounds, positive pixel anchors,
and hard-negative constraints effectively bridges the gap from coarse
localization to precise pixel-level grounding.

\section{More Experimental Details}
\label{supp:implementation}
\subsection{Datasets}
To evaluate our approach, we conduct experiments on two prominent RSVG
benchmarks supporting comprehensive evaluation for both RSREC and RSRES tasks:
\begin{itemize}
  \item \textbf{RRSIS-D} \cite{RMSIN} contains 17,402 triplets comprising RS
  images, referring expressions, and pixel-level masks. It covers diverse
  geospatial scenes across 20 semantic categories and 7 linguistic attributes.
  All images are standardized to $800\times800$ pixels with spatial resolutions
  spanning from 0.5m to 30m.
  \item \textbf{RISBench} \cite{CroBIM} is a large-scale benchmark containing
  52,472 image-text-mask triplets. It features a wide spectrum of objects
  across 26 semantic labels and 8 linguistic attributes. All images are resized
  to $512\times512$ pixels with spatial resolutions ranging from 0.1m to 30m.
\end{itemize}

\subsection{Evaluation Metrics}
Following prior works \cite{RMSIN,rsvg-zero}, we adopt three quantitative
metrics to assess performance: overall Intersection-over-Union ($oIoU$), mean
Intersection-over-Union ($mIoU$), and Precision at different thresholds
($Pr@X$). Specifically, $Pr@X$ evaluates the percentage of samples where the
IoU exceeds a specific threshold $X$, where $X\in\{0.3,0.5,0.7\}$.

\subsection{Hyperparameters and Hardware}
For representative point selection, the hyperparameters are set to
$\tau=0.1$, $\gamma=4$, and $N=2$. All experiments are conducted on a single
NVIDIA RTX3090 GPU with 24GB memory.

\begin{table}[!h]
  \caption{Ablation study on geometric prompts.}
  \label{prompts}
  \vspace{-0.3cm}
  \resizebox{0.4\textwidth}{!}{
  \begin{tabular}{ccccccc}
    \toprule
    \multirow{2.5}{*}{BBox} &
    \multirow{2.5}{*}{\makecell{Pos\\Points}} &
    \multirow{2.5}{*}{\makecell{Neg\\Points}} &
    \multicolumn{2}{c}{RSREC} & \multicolumn{2}{c}{RSRES}\\
    \cmidrule(lr){4-5}\cmidrule(lr){6-7}
    & & & Pr@0.5 & mIoU & Pr@0.5 & mIoU\\
    \midrule
    \ding{51} & & & 38.78 & 39.31 & 38.49 & 37.61\\
    & \ding{51} & & 34.21 & 34.62 & 34.88 & 33.12\\
    \ding{51} & \ding{51} & & 39.30 & 39.65 & 39.07 & 38.40\\
    \rowcolor{ourspink}
    \ding{51} & \ding{51} & \ding{51} &
    \textbf{39.82} & \textbf{40.19} & \textbf{39.82} & \textbf{39.29}\\
    \bottomrule
  \end{tabular}}

  \vspace{0.3cm}
  \caption{Ablation study on point prompt quantity.}
  \label{points}
  \vspace{-0.3cm}
  \resizebox{0.35\textwidth}{!}{
  \begin{tabular}{ccccc}
    \toprule
    \multirow{2.5}{*}{$\gamma N$, $N$} &
    \multicolumn{2}{c}{RSREC} & \multicolumn{2}{c}{RSRES}\\
    \cmidrule(lr){2-3}\cmidrule(lr){4-5}
    & Pr@0.5 & mIoU & Pr@0.5 & mIoU\\
    \midrule
    $\gamma N=8,N=1$ & 38.29 & 39.46 & 38.18 & 38.68\\
    $\gamma N=8,N=4$ & 36.08 & 37.89 & 36.37 & 37.60\\
    \midrule
    $\gamma N=2,N=2$ & 39.16 & 39.85 & 39.13 & 39.05\\
    $\gamma N=4,N=2$ & 39.19 & 39.88 & 38.98 & 39.09\\
    $\gamma N=16,N=2$ & 38.38 & 39.36 & 38.44 & 38.82\\
    \midrule
    \rowcolor{ourspink}
    $\gamma N=8,N=2$ &
    \textbf{39.82} & \textbf{40.19} & \textbf{39.82} & \textbf{39.29}\\
    \bottomrule
  \end{tabular}}
\end{table}

\begin{figure}[!h]
  \centering
  \includegraphics[width=0.95\linewidth]{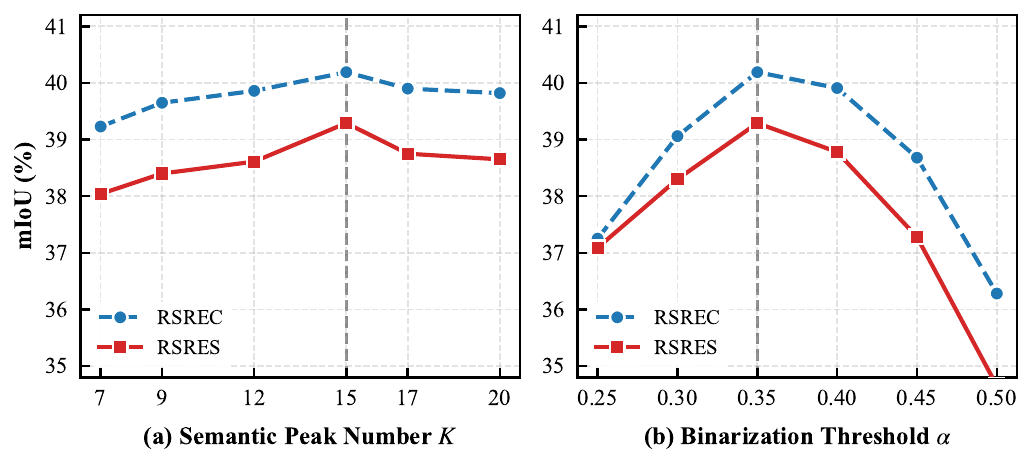}
  \vspace{-0.3cm}
  \caption{(a) Ablation on semantic peak number $K$. (b) Ablation on
  binarization threshold $\alpha$.}
  \Description{Ablation on semantic peak number and binarization threshold.}
  \label{K_alpha}
\end{figure}

\section{More Ablation Studies on RRSIS-D}
\noindent\textbf{Ablation of Various Geometric Prompts.}
Table~\ref{prompts} illustrates the combined effect of our geometric prompts.
While sparse positive points alone produce the lowest performance (34.62\%
mIoU on RSREC) due to a lack of global context, bounding boxes offer strong
spatial constraints that significantly increase the mIoU to 39.31\%. The
performance reaches its highest at 40.19\% with the full prompt combination.
This shows that while boxes and positive points help locate the target's main
structure, negative points are still essential for reducing background
distractors and clarifying boundaries in cluttered scenes.

\noindent\textbf{Ablation of Point Prompt Sampling ($N$ and $\gamma$).}
Table~\ref{points} ablates the point prompt quantity $N$ and oversampling pool
$\gamma N$, establishing $N=2$ and $\gamma N=8$ as the optimal configuration.
While a single prompt ($N=1$) cannot fully capture target structures, excessive
prompts ($N=4$) induce noisy signals. Similarly, an adequate candidate pool
($\gamma N=8$) ensures robust K-Means centroids, whereas an oversized pool
($\gamma N=16$) degrades accuracy by incorporating low-confidence outliers.
These results highlight that a concise yet representative prompt set is ideal
for training-free RSVG.

\noindent\textbf{Ablation of Semantic Peak Number ($K$).}
As shown in Figure~\ref{K_alpha}(a), we evaluate the model's performance across
different values of the semantic peak number $K$. The mIoU steadily improves
and peaks at $K=15$ for both tasks. This trend highlights a critical trade-off:
while sufficient semantic anchors ($K\leq15$) are vital for recovering the
complete physical structure of fragmented or large-scale targets, an
excessively large $K$ inadvertently incorporates background distractors,
leading to a slight performance degradation.

\noindent\textbf{Ablation of Binarization Threshold ($\alpha$).}
Figure~\ref{K_alpha}(b) examines the binarization threshold $\alpha$ for
generating the initial mask $\mathbf{M}_{pred}$, with performance reaching its
peak at $\alpha=0.35$ for both tasks. Setting $\alpha>0.40$ significantly
damages the foreground mask and cuts off genuine activations, resulting in a
sharp performance decline. Conversely, a more permissive threshold
($\alpha<0.35$) causes a milder degradation due to semantic leakage. The stable
mIoU around this peak demonstrates ExACT's robustness to threshold changes
across diverse RS scenes.

\begin{figure}[!h]
  \centering
  \includegraphics[width=\linewidth]{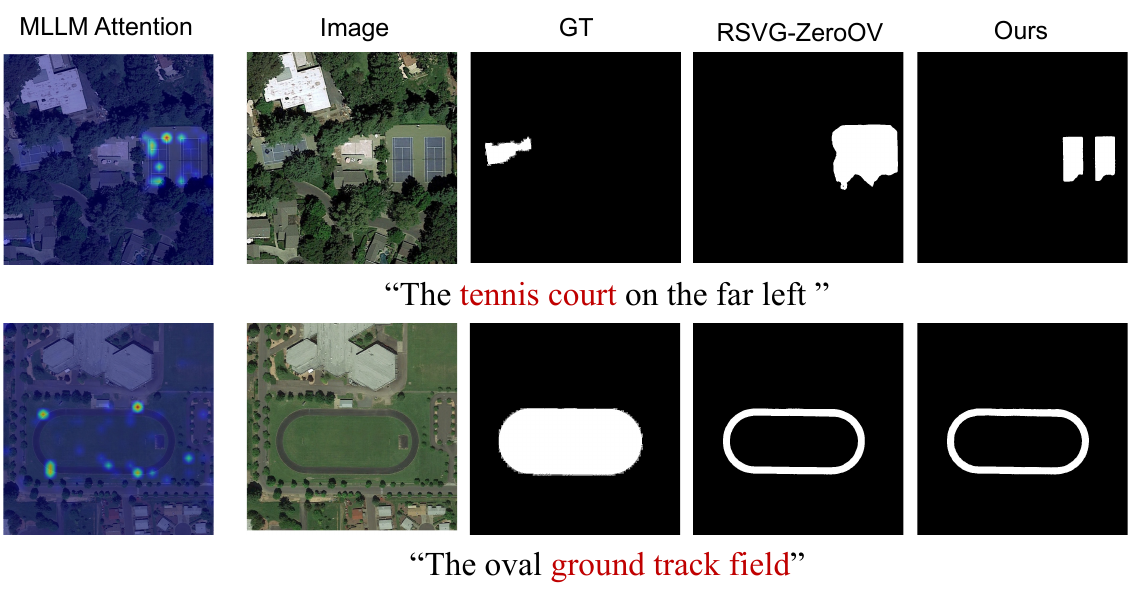}
  \caption{Failure cases of our approach on RRSIS-D dataset.}
  \vspace{-0.4cm}
  \Description{Failure cases of our approach.}
  \label{failure}
\end{figure}

\section{Limitations}
\label{supp:limitation}
Although our approach attains competitive performance in training-free RSVG,
its capacity to handle complex spatial relational logic remains limited by the
core reasoning constraints of foundational MLLMs. Specifically, while using
visual exemplars helps calibrate the MLLM's deviations in category-level visual
semantics, it still falls short in resolving spatial positional bias (e.g.,
``far left'' in the first row of Figure~\ref{failure}). Additionally, our
cascaded pipeline is inevitably limited by the inherent biases of vanilla MLLM
attention maps. In hierarchical scenarios such as the ``track field'' case in
the second row of Figure~\ref{failure}, attention activations are
disproportionately directed towards high-contrast edges due to the very strong
visual texture of the track border. As a result, this misleads the subsequent
geometric extraction to focus on exterior boundaries instead of capturing the
true semantic interior. These limitations highlight ongoing challenges for
training-free paradigms, particularly in interpreting abstract spatial logic
and bridging the hierarchical semantic-geometric gap. To overcome these issues,
future work will explore training-free attention calibration methods to address
spatial-semantic positional biases and develop causal intervention mechanisms
to reduce over-concentration on discriminative local features, thereby
broadening focus to encompass the genuine geometric structure of the queried
target.

\end{document}